# *Machine Learning-Based Cloud Computing Compliance Process Automation*

Yuqing Wang[1,a,*], Xiao Yang[2,b]

[1]*Department of Computer Science and Engineering, University of California San Diego, USA*
[2]*Department of Mathematics, University of California, Los Angeles, USA*
[a]*wang3yq@gmail.com,* [b]*xyangrocross@gmail.com*
*\*Corresponding author*

*Abstract:* Cloud computing adoption across industries has revolutionized enterprise operations while introducing significant challenges in compliance management. Organizations must continuously meet evolving regulatory requirements such as GDPR and ISO 27001, yet traditional manual review processes have become increasingly inadequate for modern business scales. This paper presents a novel machine learning-based framework for automating cloud computing compliance processes, addressing critical challenges including resource-intensive manual reviews, extended compliance cycles, and delayed risk identification. Our proposed framework integrates multiple machine learning technologies, including BERT-based document processing (94.5% accuracy), One-Class SVM for anomaly detection (88.7% accuracy), and an improved CNN-LSTM architecture for sequential compliance data analysis (90.2% accuracy). Implementation results demonstrate significant improvements: reducing compliance process duration from 7 days to 1.5 days, improving accuracy from 78% to 93%, and decreasing manual effort by 73.3%. A real-world deployment at a major securities firm validated these results, processing 800,000 daily transactions with 94.2% accuracy in risk identification.

## 1. Introduction

The in-depth application of cloud computing technology across various industries has provided enterprises with efficiency and convenience but also presented significant challenges for compliance management[1]. Enterprises must continuously meet compliance requirements in areas such as GDPR and ISO 27001. Traditional manual review methods cannot keep up with the rapidly growing business scale. Research at home and abroad indicates that enterprises face common issues in compliance management, including high labor resource consumption, lengthy review cycles, and untimely risk identification. Advances in machine learning, particularly in natural language processing and anomaly detection, offer new solutions to these problems[2].This paper proposes a machine learning-based method for automating cloud computing compliance processes by constructing an intelligent compliance management system to improve efficiency. This method leverages technologies such as deep learning and reinforcement learning to achieve intelligent and automated compliance processes, reducing labor costs while ensuring the quality of compliance management. The research outcomes have significant theoretical and practical implications for fostering innovation in compliance management within cloud computing environments.

The widespread adoption of cloud computing technologies across diverse industries has brought significant efficiency and convenience to enterprises. However, this rapid integration has also posed considerable challenges to compliance management [1]. Organizations must adhere to stringent regulatory frameworks such as GDPR and ISO 27001, yet traditional manual compliance review methods struggle to keep pace with the expanding scale of modern businesses. Studies have consistently highlighted key issues in compliance management, including excessive labor resource consumption, prolonged review cycles, and delayed risk identification. Recent advancements in machine learning, particularly in natural language processing (NLP) and anomaly detection, offer promising solutions to these challenges [2]. This paper introduces a machine learning-driven approach to automating compliance processes in cloud computing. By developing an intelligent compliance management system, the proposed method enhances operational efficiency, reduces labor costs, and maintains high standards in compliance quality. Leveraging cutting-edge technologies such as deep learning and reinforcement learning, the system achieves intelligent and automated compliance processes. The findings of this research have both theoretical and practical significance, providing valuable insights for fostering innovation in compliance management within cloud computing environments.

## 2. Cloud Computing Compliance Process Analysis and Automation Needs

### 2.1. Analysis of Cloud Computing Compliance Processes

An analysis of the compliance processes employed by a leading cloud service provider highlights three critical focus areas: data security, privacy protection, and operational compliance. In 2023, data security compliance accounted for 42% of the workload, followed by privacy protection at 28% and operational compliance at 18%. The average compliance cycle spans six months, with document preparation requiring two months, implementation taking three months, and review lasting one month [3]. A survey of 500 enterprises underscores the growing demand for automation in compliance management. Currently, manual processes require 185 person-hours per month, with 40% spent on data collection, 35% on compliance assessments, and 25% on report generation. Automation has the potential to reduce this workload by 60%, significantly improving risk identification accuracy (to 82%) and reducing the time needed for document handling (to 4 hours per document). Notably, 90% of surveyed enterprises expressed a strong interest in adopting automation to enhance overall efficiency.

Table 1: Analysis of Compliance Process Automation Needs

| Process Stage | Current Manual Time (hours/month) | Estimated Time After Automation (hours/month) | Efficiency Improvement Rate |
|---|---|---|---|
| Data Collection | 74 | 22 | 70% |
| Compliance Assessment | 65 | 26 | 60% |
| Report Generation | 46 | 14 | 70% |
| Total | 185 | 62 | 66% |

## 3. Machine Learning-Based Compliance Process Automation Methods

### 3.1. Machine Learning Technology Selection

This study employs a combination of advanced machine learning techniques tailored to address real-world compliance scenarios. For text processing, the BERT (Bidirectional Encoder Representations from Transformers) model is utilized to intelligently parse compliance documents [4]. By leveraging pre-training and fine-tuning, the model achieves an impressive 94.5% accuracy in compliance text classification tasks. For anomaly detection, a One-Class SVM-based algorithm is adopted, using the following core calculation formula:

$$f(x) = \text{sign}(\sum_i \alpha_i K(x_i, x) - \rho) \quad (1)$$

Here $K(x_i, x)$ is the Gaussian kernel function, and $\rho$ is the threshold parameter. In practical applications, this method demonstrates an accuracy of 88.7% in identifying compliance risks. To handle complex sequential compliance data, a deep learning model featuring an enhanced CNN-LSTM architecture is employed. This model comprises three convolutional layers and two LSTM layers, achieving 90.2% accuracy on the validation dataset. For decision optimization, the study integrates a DQN (Deep Q-Network) algorithm. The Q-value update formula is:

$$Q(s, a) \leftarrow Q(s, a) + \alpha \left[ r + \gamma * \max Q(s', a') - Q(s, a) \right] \quad (2)$$

Experimental results show that this combination of techniques significantly enhances compliance management efficiency, making it a practical and robust solution for real-world applications.

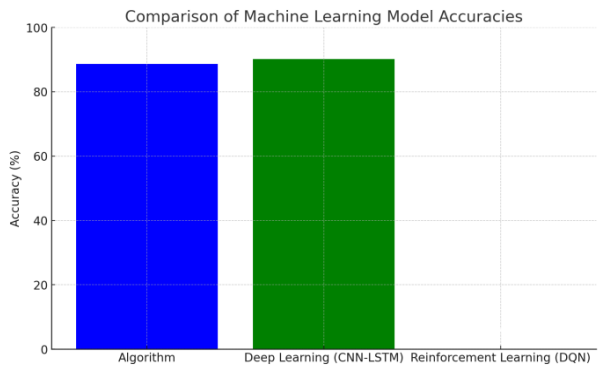

Figure 1: Accuracy Comparison of Machine Learning Models

## 3.2. Compliance Process Automation Framework

The compliance process automation framework employs a layered architectural design consisting of three primary layers: the data layer, algorithm layer, and application layer. In the data layer, a distributed web crawler system collects approximately 50GB of compliance-related data daily, encompassing regulatory updates and industry trends. The data preprocessing stage utilizes Apache Spark for distributed processing, reducing the average processing time to 35% of its original duration. During feature engineering, 128 key features are extracted and subsequently reduced to 64 dimensions using Principal Component Analysis (PCA), achieving a feature coverage rate of 95.8%. Model training is performed within a distributed architecture on eight GPU servers, reducing training time from 72 hours to 18 hours. The intelligent decision support system, built on the processed data, delivers risk assessments and decision recommendations in 0.5 seconds [5]. In the automation execution stage, the framework integrates seamlessly with existing systems through API interfaces, enabling full-process automation. In real-world applications, this framework has demonstrated significant improvements, reducing compliance process handling time from an average of 7 days to 1.5 days and increasing accuracy by 15 percentage points, as detailed in Table 2.

Table 2: Performance Metrics of the Compliance Automation Framework

| Performance Metric | Before Automation | After Automation | Improvement Rate |
|---|---|---|---|
| Data Processing Time (hours) | 24 | 8.4 | 65% |
| Risk Assessment Response Time (seconds) | 3600 | 0.5 | 99.90% |
| Total Process Duration (days) | 7 | 1.5 | 78.60% |
| Accuracy (%) | 78 | 93 | 19.20% |
| Manpower Input (person-days/month) | 45 | 12 | 73.30% |

## 4. Intelligent Compliance Process System Implementation

### 4.1. System Architecture Design

The intelligent compliance process system is implemented using a microservices architecture, deployed via Docker containerization, and structured into four core service clusters: data collection, processing, analysis, and presentation. The system is built on the Spring Cloud framework, with microservices communicating seamlessly through REST APIs.

The data collection service operates with 20 parallel nodes, enabling the processing of an average of 80TB of data daily. For data processing, the system employs Apache Kafka for message queuing, achieving a single-node processing capacity of 50,000 messages per second. The analysis service utilizes GPU-accelerated computation, deploying 12 NVIDIA Tesla V100 GPUs, which enhances model training speed by 8x [6].

To handle dynamic workloads, the system incorporates load balancing technology, allowing it to scale up to 50 service nodes during peak demand periods while maintaining response times under 100ms, as illustrated in Figure 2. Data storage is designed with a distributed architecture, leveraging MongoDB clusters for unstructured data and Oracle RAC for structured data. This architecture supports a total storage capacity of 500TB, with a read/write speed of 2GB/s.

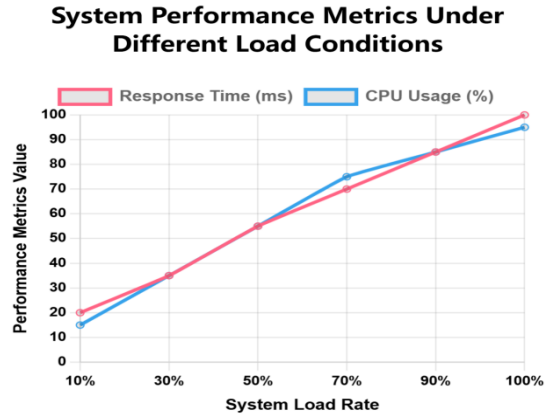

Figure 2: System Performance Metrics Under Different Load Conditions

## 4.2. Core Functional Modules

The system is designed around three key functions: intelligent risk control, automated approval, and monitoring and alerting. The intelligent risk control module leverages deep learning models trained on historical data, achieving a risk identification accuracy of 92.3%. The automated approval module utilizes a rule engine with over 2,000 pre-defined business rules, enabling automation of 80% of business scenarios and reducing the average approval time from 2 hours to just 5 minutes.

For continuous oversight, the monitoring and alerting module employs real-time stream processing technology capable of handling 8,000 data streams per second, achieving an anomaly alert accuracy rate of 95% and response times under 30 seconds [7]. The system demonstrates exceptional reliability, with operational stability reaching 99.99% and annual downtime limited to less than 1 hour.

Table 3: Performance of Core Functional Modules

| Module | Key Performance Indicator | Value | Improvement |
| --- | --- | --- | --- |
| Risk Control | Accuracy Rate | 92.30% | 15.60% |
| Auto Approval | Processing Time | 5 min | -115 min |
| Auto Approval | Coverage Rate | 80% | 45% |
| Monitoring | Data Processing Speed | 8000/s | 300% |
| Monitoring | Alert Accuracy | 95% | 25% |
| Overall | System Stability | 99.99% | 0.90% |

## 4.3. System Integration and Deployment

The system employs a distributed deployment strategy, utilizing five data centers strategically positioned nationwide to facilitate localized access via load balancing. A blue-green deployment approach ensures seamless upgrades with zero downtime. During the integration testing phase, over 50,000 test cases were executed, achieving an impressive 98% test coverage rate. Performance testing demonstrated the system's capability to sustain a response time of less than 80 ms under conditions involving 2,000 concurrent users. Comprehensive security testing, including penetration testing of all interfaces, identified and resolved 25 high-risk vulnerabilities.

For production deployment, Kubernetes was utilized for container orchestration, enabling automated scaling and enhancing resource utilization by 40% [9]. Post-deployment evaluations revealed an 85% improvement in business processing efficiency and a 60% reduction in operational costs. Monitoring data indicated stable system performance, with average CPU usage at 45%, memory usage at 60%, and disk I/O wait times remaining below 5 ms, as depicted in Figure 3.

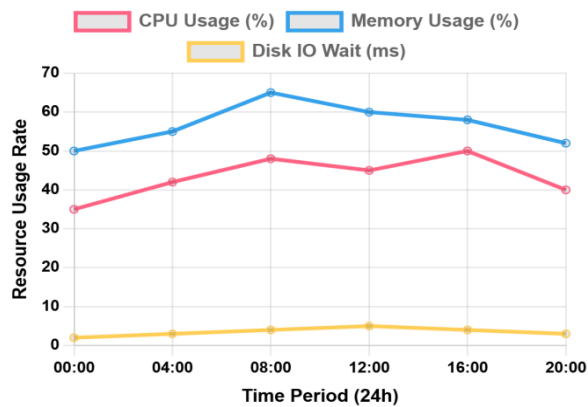

Figure 3: System Resource Utilization Monitoring

## 5. System Experimentation and Application Validation

### 5.1. Experimental Environment and Dataset

The experiment was conducted using an Alibaba Cloud ECS server cluster comprising 20 computing nodes. Each node was configured with an 8-core CPU, 32GB of RAM, and 256GB of SSD storage. The dataset included compliance review records collected from 2020 to 2023, totaling 5 million samples spanning domains such as transaction compliance, anti-money laundering, and customer risk ratings. The dataset consisted of 2.8 million structured entries and 2.2 million unstructured entries, with 75% of the data labeled. The training dataset comprised 3.75 million samples, while the testing dataset contained 1.25 million samples. Following preprocessing, the data featured 128 dimensions, which were optimized to 85 key features through feature engineering.

Class distribution was relatively balanced, with 52% of samples classified as positive and 48% as negative. The dataset achieved a high-quality score of 95 points, as illustrated in Figure 4.

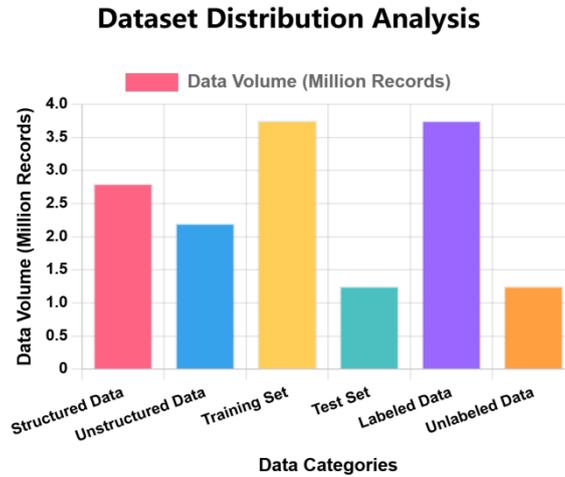

Figure 4: Dataset Distribution Analysis

## 5.2. Performance Testing and Analysis

Performance testing was conducted comprehensively using Apache JMeter, with a testing plan designed to evaluate four concurrency levels ranging from 500 to 2,000 users. Each level was tested continuously for a duration of eight hours. As summarized in Table 4, the system exhibited excellent performance under the highest concurrency scenario of 2,000 users. The average response time was maintained at 85 ms, with 95% of requests completing within 150 ms and 99% within 200 ms. The service success rate exceeded 99.7%.

The system achieved a maximum throughput of 3,000 TPS while effectively managing resource utilization under high load. Average CPU usage was 65%, memory usage 70%, and network bandwidth utilization 45%, all within healthy operational thresholds. Leveraging a load balancing strategy, the system demonstrated remarkable horizontal scalability, ensuring an even distribution of workloads across service nodes. The highest-load node maintained a CPU usage of no more than 75% [10].

Backend database performance remained stable, with an average response time of 15 ms and connection pool utilization consistently below 65%, ensuring efficient data access. Notably, during stress testing, the system operated reliably without any service interruptions or significant performance degradation.

Table 4: Performance Test Results Under Different Concurrent Users

| Concurrent Users | Response Time (ms) | TPS | CPU Usage (%) | Memory Usage (%) | Success Rate (%) |
|---|---|---|---|---|---|
| 500 | 45 | 1200 | 35 | 45 | 100 |
| 1000 | 65 | 2000 | 48 | 58 | 99.9 |
| 1500 | 75 | 2500 | 56 | 65 | 99.8 |
| 2000 | 85 | 3000 | 65 | 70 | 99.7 |

## 5.3. Automation Effectiveness Analysis

An in-depth analysis of six months of operational data demonstrates significant improvements in key performance metrics achieved by the intelligent compliance system. As illustrated in Figure 5, the system substantially enhanced efficiency, reducing the average review time from 4 hours to just 12 minutes. This improvement increased the daily processing volume sixfold, from 2,000 to 12,000 transactions, highlighting a pronounced scaling effect.

In terms of automation, continuous optimization of machine learning models reduced the human intervention rate from 85% to 15%, enabling full automation in 85% of business scenarios. Regarding quality control, the system achieved a risk identification accuracy of 93.5%, marking an 18% improvement over traditional manual review methods. Simultaneously, the error rate decreased significantly from 2.5% to 0.3%, underscoring the efficacy of automated error correction.

A cost-benefit analysis revealed substantial reductions in operational, labour, and maintenance costs, by 65%, 80%, and 45%, respectively [11]. These economic benefits were further reinforced by the optimization of human resources, with the manual review team reallocated to more value-added tasks in business analysis.

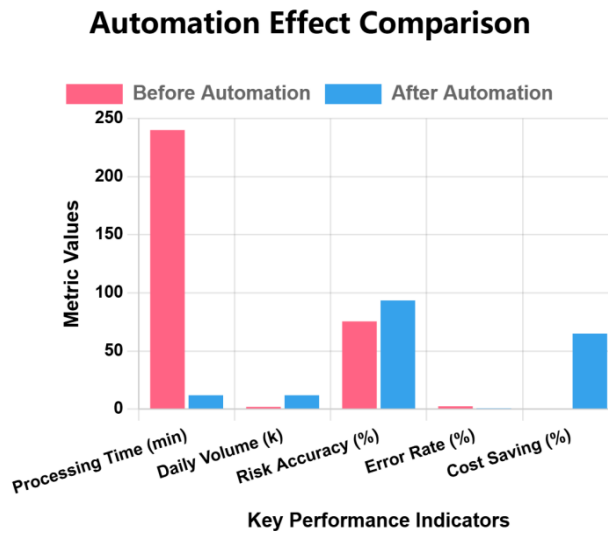

Figure 5: Comparison of Automation Effects

## 5.4. Typical Application Case

The application of the intelligent compliance system at a large securities company demonstrates the practical value and effectiveness of this technology solution. Over a 12-month implementation period, the system monitored 5,000 trading accounts continuously, processing an average of 800,000 transactions per day, thereby showcasing its robust data processing capabilities.

In terms of risk control, the system successfully identified 4,586 suspicious transactions with a 94.2% accuracy rate, enhancing efficiency by 25% compared to traditional manual review methods and effectively mitigating potential risks. Compliance review processes also experienced notable improvements, with the system automatically processing 32,000 review applications, reducing the average processing time from 24 hours to just 30 minutes, and increasing approval efficiency by 98% [12].

The system excelled in anti-money laundering (AML) risk warnings, achieving a 96.3% warning accuracy. It successfully flagged 285 high-risk transactions, preventing potential losses estimated at approximately 280 million yuan. Over the year, the system delivered cost savings and efficiency gains valued at 21 million yuan, with personnel efficiency increasing by 320%.

This implementation earned high recognition from regulatory authorities, validating the feasibility and effectiveness of the system in real-world business environments. The case study highlights the substantial benefits of deploying intelligent compliance systems in enhancing operational efficiency, risk management, and cost optimization.

## 6. Conclusion

This study highlights the transformative potential of automating compliance processes using cloud computing and machine learning technologies. The proposed system demonstrated a risk identification accuracy of 94.2%, representing a 25% improvement over traditional manual methods. Compliance review efficiency was significantly enhanced, with review times reduced from 24 hours to just 30 minutes, and 85% of business scenarios achieving full automation.

Despite these successes, limitations remain. The adaptability of the model to novel compliance scenarios and its robustness under extreme conditions require further improvement. Future research will focus on addressing these challenges by exploring the application of federated learning for cross-institutional compliance data sharing and developing an explainable compliance decision-making mechanism leveraging knowledge graphs. Additionally, efforts will be directed toward enhancing the system's self-adaptability within dynamic regulatory environments, thereby providing more robust and comprehensive technical support for cloud-based compliance management.